\documentclass[10pt,twocolumn,letterpaper]{article}

\usepackage{3dv}
\usepackage{times}
\usepackage{epsfig}
\usepackage{graphicx}
\usepackage{amsmath}
\usepackage{amssymb}

\usepackage{caption}
\usepackage{multirow}
\usepackage{verbatim}

\usepackage[pagebackref=true,breaklinks=true,letterpaper=true,colorlinks,bookmarks=false]{hyperref}

\threedvfinalcopy

\ifthreedvfinal\fi
\begin{document}

\title{DeepC-MVS: Deep Confidence Prediction for Multi-View Stereo Reconstruction}

\author{Andreas Kuhn\textsuperscript{1}, Christian Sormann\textsuperscript{2}, Mattia Rossi\textsuperscript{1,3}, Oliver Erdler\textsuperscript{1}, Friedrich Fraundorfer\textsuperscript{2} \\ \\
\textsuperscript{1} Sony Europe B.V., \textsuperscript{2} Graz University of Technology, \textsuperscript{3} \'Ecole Polytechnique F\'ed\'erale de Lausanne}

\definecolor{dgreen}{rgb}{0.0, 0.5, 0.0}

\twocolumn[{
\renewcommand\twocolumn[1][]{#1}
\maketitle
\vspace{-1.0cm}
\begin{center}
    \centering
    \includegraphics[width=0.245\textwidth]{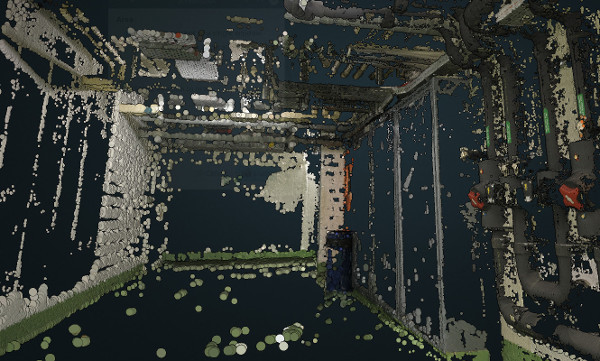}
    \includegraphics[width=0.245\textwidth]{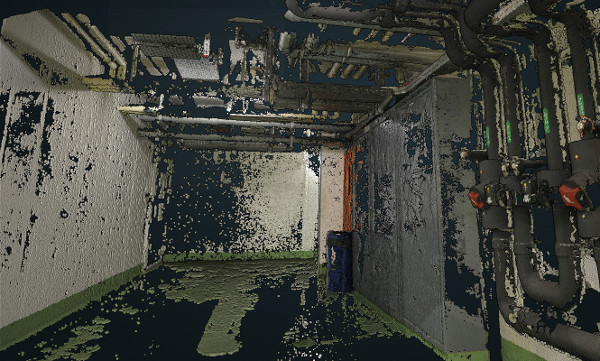}
    \includegraphics[width=0.245\textwidth]{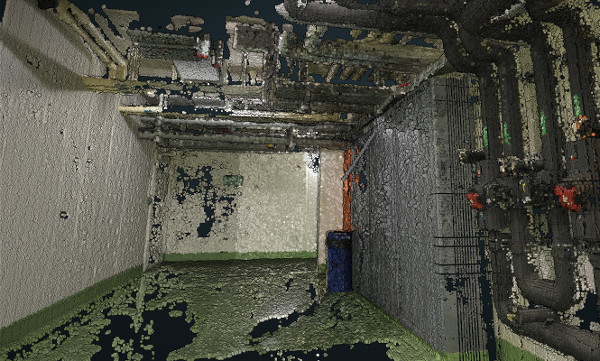}
    \includegraphics[width=0.245\textwidth]{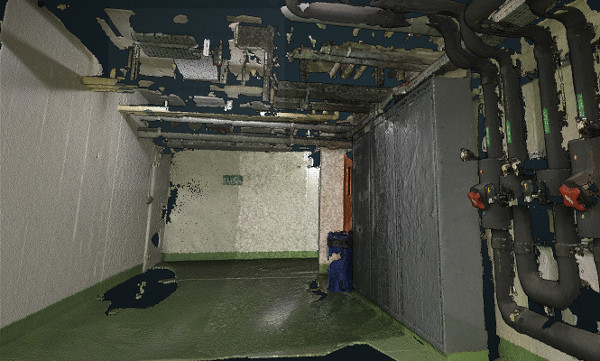}
    \vspace{-6pt}
    \captionof{figure}{
    \texttt{pipes} point clouds from the ETH3D webpage \cite{schoeps2017cvpr}.
    Left to right: COLMAP \cite{schoenberger:2016mvs}, ACMM \cite{Xu:arxiv:2019} and our pipelines DeepC-MVS$_{\text{fast}}$ and DeepC-MVS, employing deep-confidence-based filtering and refinement, respectively.}
\end{center}
}]

\begin{abstract}
Deep Neural Networks (DNNs) have the potential to improve the quality of image-based 3D reconstructions.
However, the use of DNNs in the context of 3D reconstruction from large and high-resolution image datasets is still an open challenge, due to memory and computational constraints.
We propose a pipeline which takes advantage of DNNs to improve the quality of 3D reconstructions while being able to handle large and high-resolution datasets.
In particular, we propose a confidence prediction network explicitly tailored for Multi-View Stereo (MVS) and we use it for both depth map outlier filtering and depth map refinement within our pipeline, in order to improve the quality of the final 3D reconstructions.
We train our confidence prediction network on (semi-)dense ground truth depth maps from publicly available real world MVS datasets.
With extensive experiments on popular benchmarks, we show that our overall pipeline can produce state-of-the-art 3D reconstructions, both qualitatively and quantitatively.
\end{abstract}

\section{Introduction}

\begin{figure*}[!t]
\centering
\includegraphics[width=0.97\textwidth]{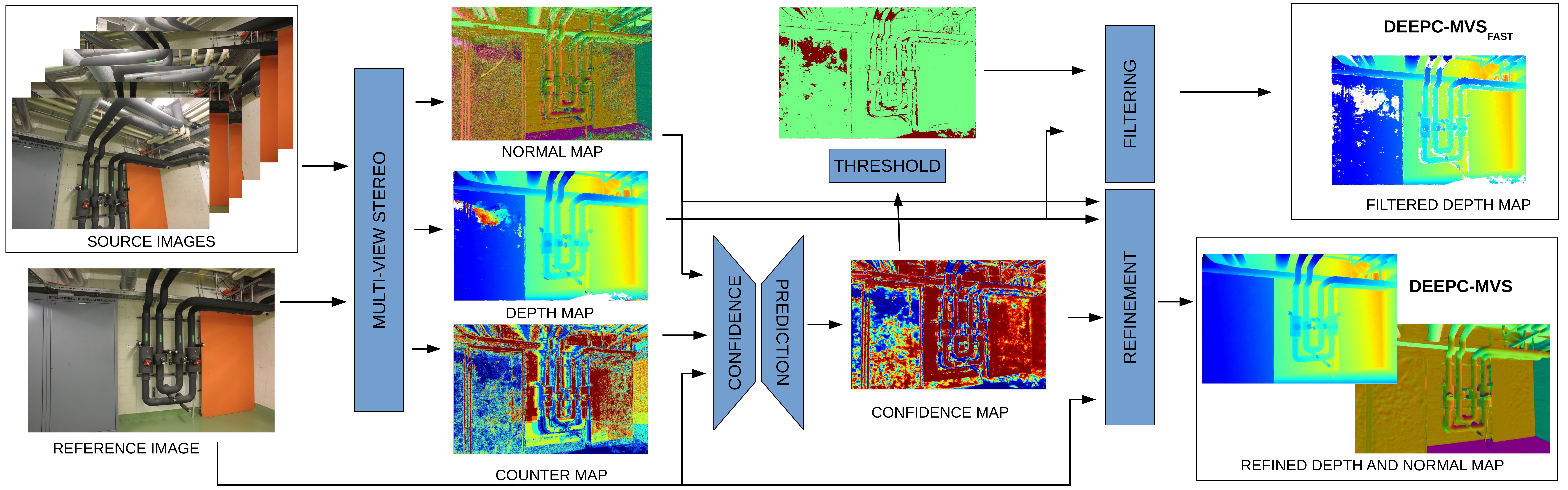}
\vspace{-6pt}
\caption{Our proposed confidence-based filtering $\text{DeepC-MVS}_{\text{fast}}$ (top) and refinement DeepC-MVS (bottom) providing improved accuracy and completeness.}
\label{fig:pipeline}
\end{figure*}

Multi-View Stereo (MVS) methods permit the 3D reconstruction of a scene from multiple images for which the inner and outer camera calibration is known.
The major challenge is the reconstruction of 3D point clouds as complete as possible while minimizing the number of outliers and maintaining a high accuracy of the points.
At the basis of MVS methods is the estimation of depth maps from input images, therefore the completeness and accuracy of the final point cloud depends inevitably on the completeness and accuracy of the computed depth maps.

A possible approach to minimize outliers is to use filtering methods, which preserve most accurate measurements in depth maps and remove unreliable ones.
Unfortunately, this typically results in rather sparse point clouds, which is not desirable for applications involving photo-realistic 3D rendering, as these have a need for complete models.
In order to improve the completeness of the reconstructed 3D scenes, regularization techniques are typically useful.
Many conventional MVS methods regularize the 3D matching cost volumes \cite{Kolmogorov:2002,Sun:2003,Hirschmuller2008} in a global optimization fashion.
The main drawback of these techniques are the high memory and computational requirements, due to the large size of the cost volume itself, which in practice prevents their applicability to high resolution datasets \cite{kuhn:gcpr:2019}.
The 3D cost volumes can also be fed to a Deep Neural Network (DNN) \cite{huang:cvpr:2018,yao:eccv:2018,Luo_2019_ICCV,DBLP:journals/corr/abs-1912-06378}  which optimizes the volume by means of a learned regularization.
However, the high memory requirements of DNNs in these approaches make them rather unsuitable for processing high-resolution image datasets.

For the efficient estimation of depth maps, the PatchMatch (PM) method \cite{bleyer:bmvc:2011} has demonstrated high quality results without the need to handle a global cost volume.
In fact, PM adopts a stochastic search over the depth space.
Efficient implementations exist, which further optimize the computational efficiency by proposing parallelization schemes \cite{Galliani_2015_ICCV,zheng:cvpr:2014} and sophisticated view selection approaches \cite{schoenberger:2016mvs}.
However, due to the local optimization adopted in PM, the results lack in completeness, hence further processing is needed.

In this paper our contribution is twofold.
First, we extend the multi-scale version of PM proposed in \cite{Xu:arxiv:2019}, which results in depth maps characterized by fewer outliers and larger completeness.
Second, and our main contribution, we present
a confidence prediction network explicitly tailored for MVS data and use the resulting confidence maps to guide two applications: outlier filtering on the computed depth maps and their refinement.
We carry out extensive experiments on popular benchmarks and show that our two pipelines can produce state-of-the-art 3D reconstructions, both qualitatively and quantitatively.

\section{Related Work}

In this section we recall the main MVS and Confidence prediction works relevant to our proposed pipeline.

\textbf{Multi-View Stereo}:
In a similar approach to classical two-view stereo methods, which build a cost volume by matching image patches along the epipolar lines, plane-sweep-based MVS methods construct a cost volume for a set of given plane hypothesis \cite{Gallup:cvpr:planesweep}.
Instead of the number of disparities, the depth of the volume is defined by the number of planes.
As this leads to a significant consumption of computational resources, Bleyer et~al. \cite{bleyer:bmvc:2011} propose the PM algorithm~\cite{Barnes:siggraph:patchmatch}, which tries to reduce the amount of computed matching costs by propagating depth hypothesis across the image.
This strategy has also been implemented for the MVS case \cite{bailer:eccv:2012}.
Zheng et~al. \cite{zheng:cvpr:2014} make use of a probabilistic scheme for view selection in PM MVS~\cite{bailer:eccv:2012}, which is improved upon by Sch\"{o}nberger et~al. \cite{schoenberger:2016mvs}.
In order to increase the completeness of depth estimates, Romanoni and Matteucci \cite{Romanoni:iccv:2019} introduce a method which propagates the depth from local planes estimated on superpixels.
This approach is extended upon by Kuhn~et~al.~\cite{kuhn:gcpr:2019}, who propose a region growing for the superpixels and additional outlier filtering strategies.
A black-red checkerboard sampling scheme was used by Galliani~et~al. \cite{Galliani_2015_ICCV} in order to decrease the runtime of {PM-based} MVS, which was further improved upon by the multi-scale approach of Xu and Tao \cite{Xu:arxiv:2019}.

In recent years, DNN-based approaches to MVS~\cite{huang:cvpr:2018,yao:eccv:2018,Xue2019MVSCRFLM}, working with cost volumes, have been established.
Yao et~al. \cite{yao:cvpr:2019} have extended their approach to process higher resolution imagery.
However, as described in \cite{yao:cvpr:2019}, this method is not able to process resolutions as high as the ones in the ETH3D high-res benchmark \cite{schoeps2017cvpr}.
For this reason, we make use of the PM-based approach, which bypasses the processing of a large cost volume for high resolution imagery.
Furthermore, we adopt the multi-scale approach of \cite{Xu:arxiv:2019}, to increase the robustness of the method when processing images with large amounts of {untextured} regions.

\textbf{Confidence Prediction:}
The estimation of a depth map confidences is a key component in 3D reconstruction.
A confidence map can be calculated as the local patch comparison cost, based on metrics like the Normalized Cross Correlation.
A detailed analysis of the chosen metric influence on the local matching costs is given by Hirschm\"{u}ller et al. \cite{Hirschmuller2008_2} and by Hu et al. \cite{Hu:2012:QEC:2377349.2377545}.
Moreover, the confidence can be derived from a globally optimized cost volume within the cost aggregation process \cite{Seitz2006}.
The latter allows the consideration of global cost terms like the overall smoothness of a disparity map  \cite{Kolmogorov:2002,Sun:2003,Hirschmuller2008}.

It has been shown that machine learning methods improve the quality of the confidence prediction, e.g, by employing hand-crafted features as input for a random forest classifier \cite{Haeusler:2013,Spyropoulos:cvpr:2016,poggi:2016,Park:2015}.
The use of automatically learned features for confidence prediction was firstly proposed by Seki et al.~\cite{Seki:2016}, while Poggi et al.~\cite{poggi:bmvc:2017} proposed the first end-to-end trained confidence prediction network fed with the sole disparity map.
Further improvements are possible by exploiting local consistencies \cite{poggi:2017}, adding the image as additional input to the network \cite{tosi:2018} or using extended information from the entire cost volume as input \cite{Kim_2019_CVPR}. 
Due to scalability issues, we avoid processing on global cost volumes and focus on methods that can be applied in the image domain.
Since all confidence prediction approaches working in the image domain are limited to the stereo scenario, in this article we propose the first confidence prediction network specifically designed for MVS-derived depth maps.

\section{Algorithm}

In this section we first provide a detailed explanation of our proposed MVS algorithm based on PM \cite{Barnes:siggraph:patchmatch}. Further, we present our confidence prediction, which estimates confidence maps for corresponding MVS depth maps.

\subsection{Multi-View Stereo} \label{subsec:mvs_algorithm_descr}

\begin{figure}[!ht]
\centering
\includegraphics[width=0.21\textwidth]{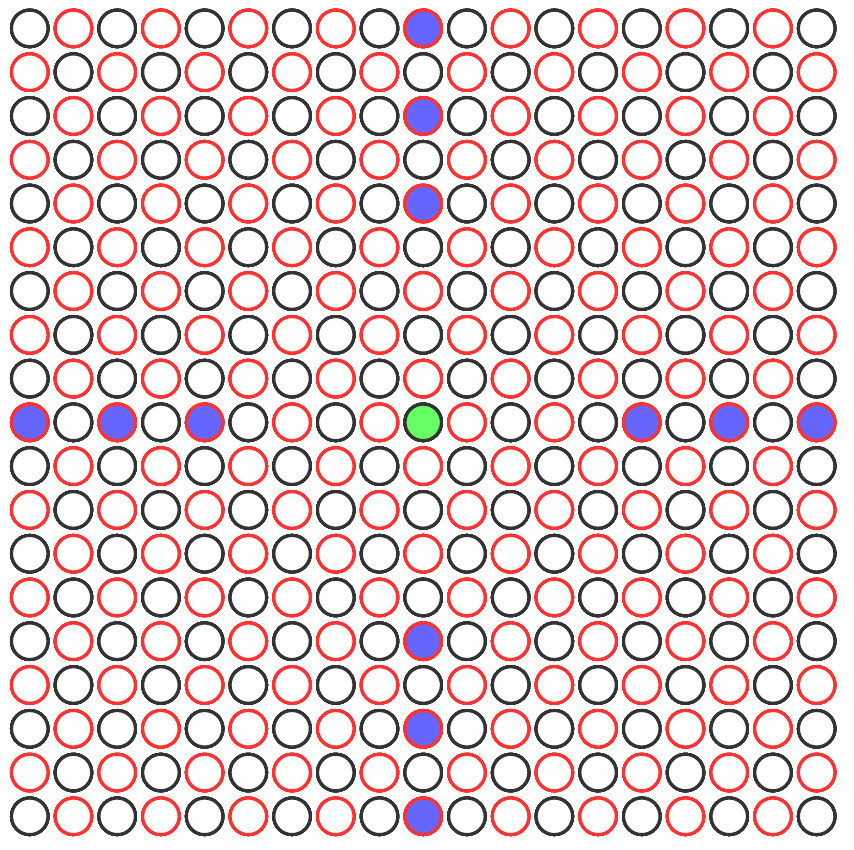}
\hspace{0.5cm}
\includegraphics[width=0.21\textwidth]{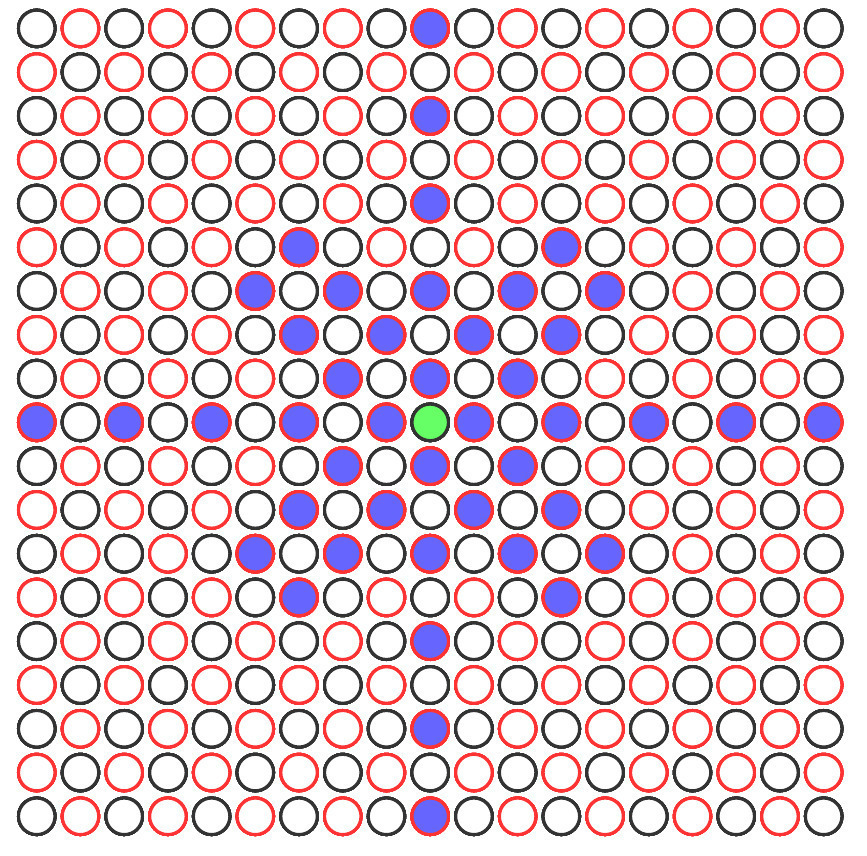}
\vspace{-6pt}
\caption{
Our checkerboard sampling pattern (left) and the ACMM \cite{Xu:arxiv:2019} one (right), with the current pixel in green and the drawn samples in blue.
For a concise visualization, the number of samples has been adapted to match the grid size: three samples per direction are represented, instead of six.
}
\label{fig:sampling_pattern}
\end{figure}

Our MVS pipeline employs the PM \cite{Barnes:siggraph:patchmatch} paradigm with a red-black checkerboard propagation scheme \cite{Galliani_2015_ICCV} and the bilateral weighted cross correlation \cite{schoenberger:2016mvs} as the photometric matching cost. In our PatchMatch~\cite{Barnes:siggraph:patchmatch} implementation we draw from six potential samples in each direction, yielding a total of $24$ hypotheses, as we found that sampling from more hypotheses increases the chance of including outliers with low matching costs.
The proposed sampling pattern is depicted in Fig.~\ref{fig:sampling_pattern}.
We remove samples close to the central pixel, following the intuition that these hypotheses are covered by the employed perturbation scheme. We perturb the depth and normal estimates according to the scheme presented in \cite{schoenberger:2016mvs}, where the parameter $\epsilon$ is calculated as $\epsilon = 2^{-i}$. In this context, the variable $i$ denotes the current red-black iteration.
Similar to \cite{Xu:arxiv:2019}, we also test hypothesis which result from combinations of the current, perturbed and random depth estimates with their respective normal vectors. This is done during the red and black sub-steps of each iteration.

Furthermore, we incorporate a plane-based depth propagation strategy \cite{schoenberger:2016mvs}.
In particular, rather than directly using the depth estimate at a given sampling location as a new hypothesis, we make use of the local plane defined by the depth and normal estimate at this location.
By means of intersecting the viewing ray of the destination pixel with the local plane defined at the sampling location, we are able to propagate rapid changes in depth values more effectively. Following \cite{Xu:arxiv:2019}, we also adopt a three level coarse to fine estimation scheme and include a geometric consistency term and detail restorer on three hierarchy levels with a down sampling factor of $0.5$.

Finally, concerning view selection, we adapt the scheme proposed by Xu and Tao \cite{Xu:arxiv:2019}, which selects the eight best hypothesis candidates from the sampling pattern based on their respective matching costs. We increase the number of candidates considered when updating the current estimate to $16$, as we want to minimize the influence of outliers in this step. We visualize example depth maps from our MVS algorithm in Fig.~\ref{fig:synth}.

\subsection{Confidence Prediction} \label{sec:confpred}

\begin{figure*}[!t]
\centering
\includegraphics[width=0.24\textwidth]{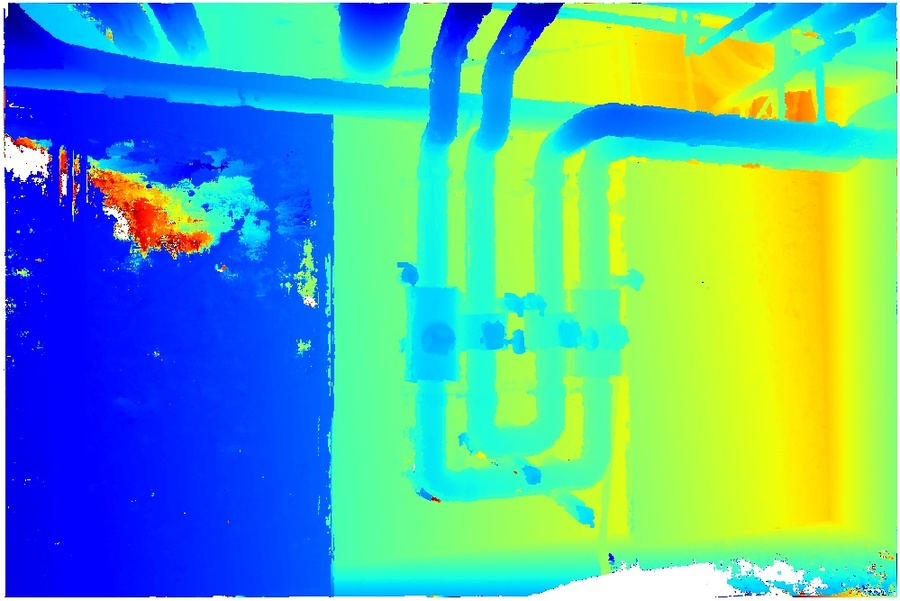} \,
\includegraphics[width=0.24\textwidth]{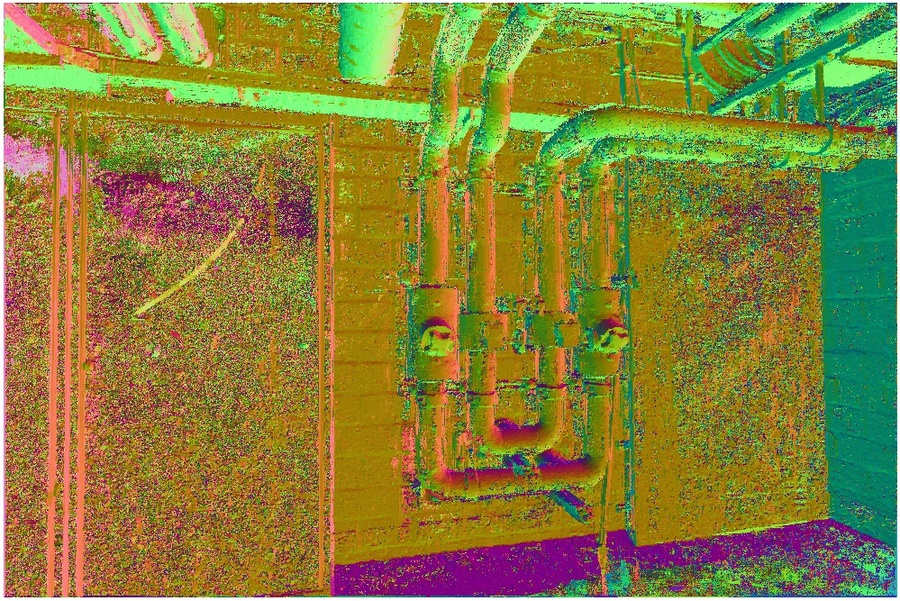} \,
\includegraphics[width=0.24\textwidth]{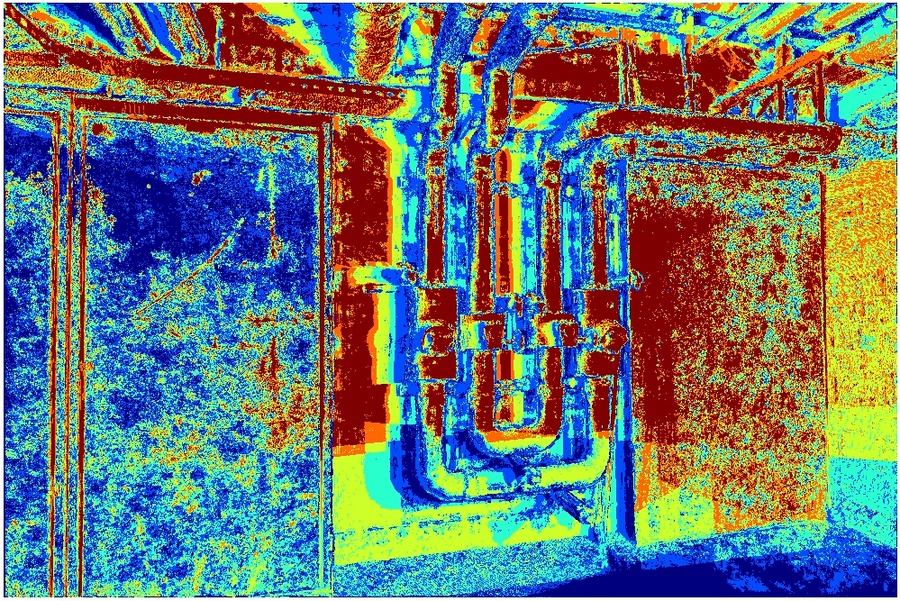} \,
\includegraphics[width=0.24\textwidth]{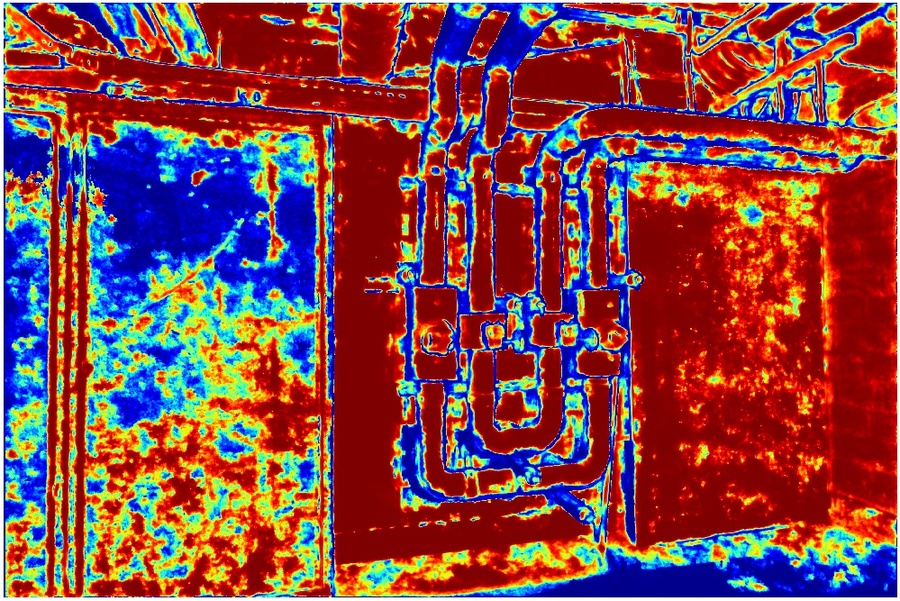}\\
\vspace{-6pt}
\caption{
From left to right, the estimated depth, normal, counter map and our predicted confidence.
Beside the RGB image, we feed the network with the normal and the counter maps.
The latter is color coded such that pixels whose depth is verified by 0 source images are in  blue and by 5 images are in red.
The confidence map pixels are color coded from blue (low confidence) to red (high confidence).}
\label{fig:normalcounter}
\end{figure*}

\begin{figure}[!t]
\centering
\setlength{\tabcolsep}{0.5pt}
\renewcommand{\arraystretch}{0.1}
\begin{tabular}{cccccc}
\includegraphics[width=0.14\textwidth]{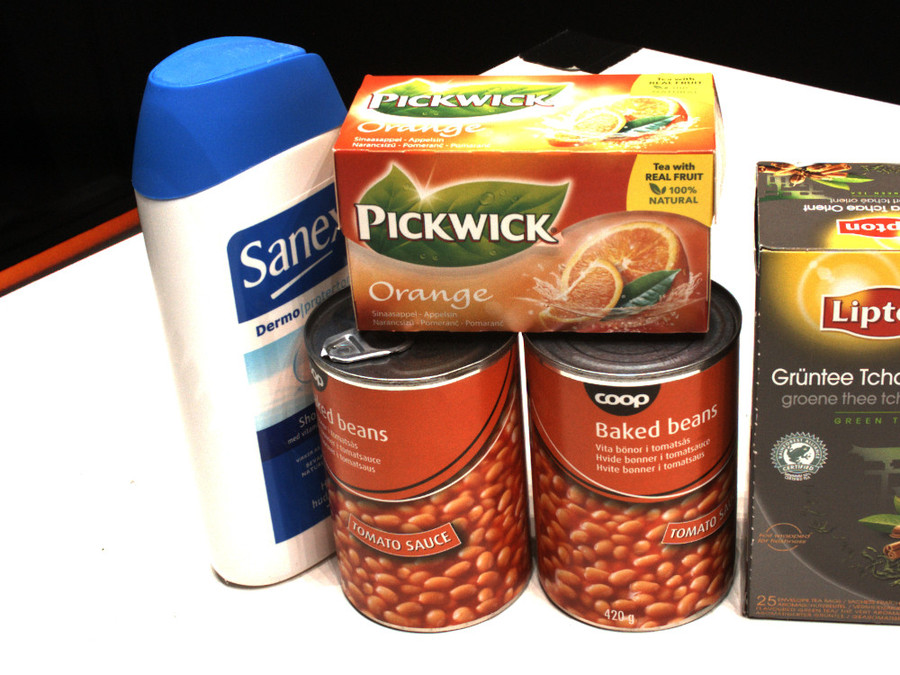} &
\includegraphics[width=0.16\textwidth]{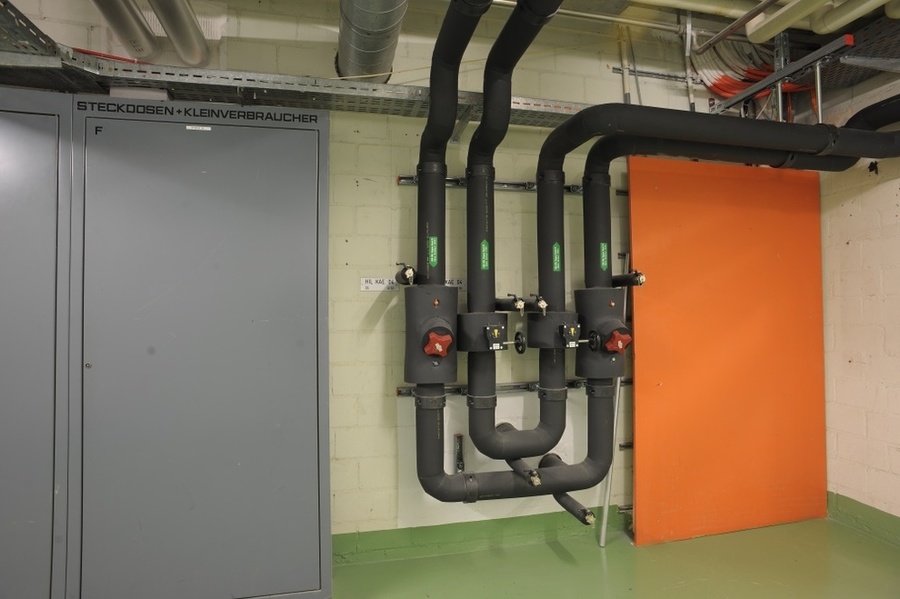} &
\includegraphics[width=0.16\textwidth]{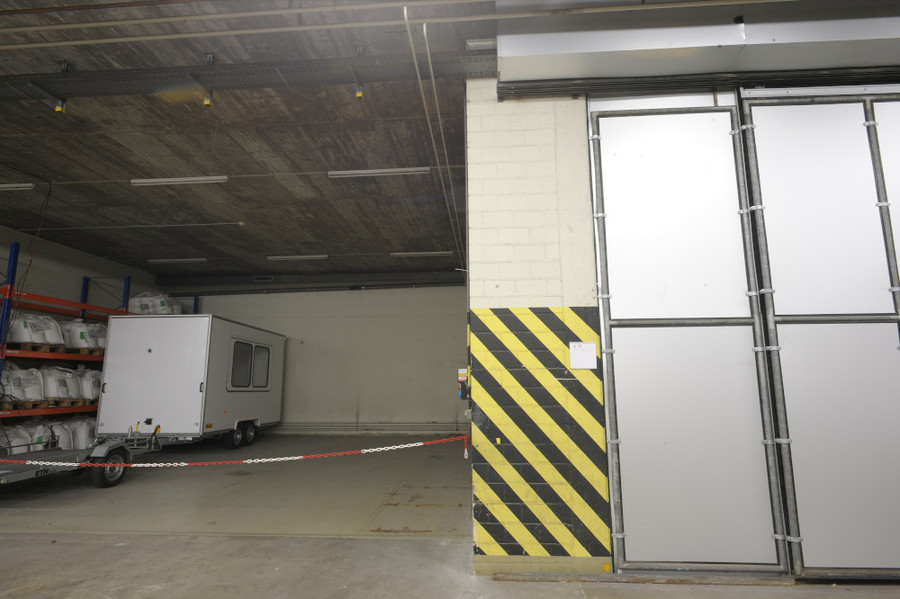} \\[0.5pt]
\includegraphics[width=0.14\textwidth]{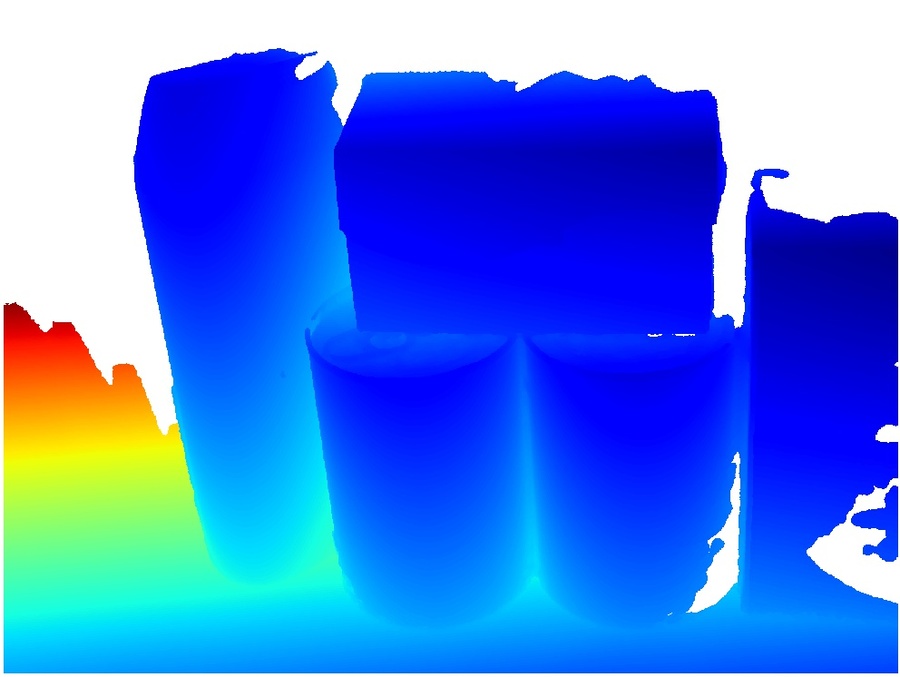} &
\includegraphics[width=0.16\textwidth]{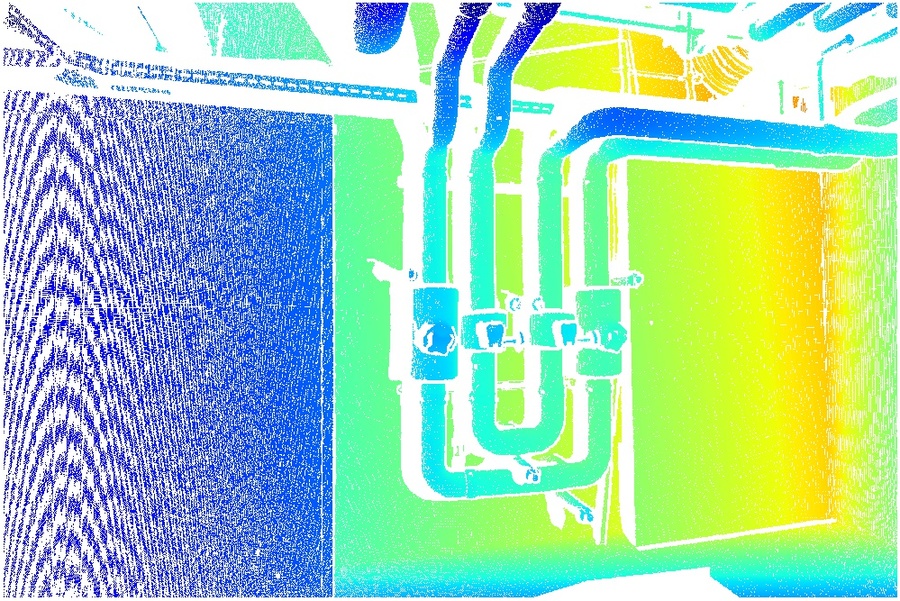} &
\includegraphics[width=0.16\textwidth]{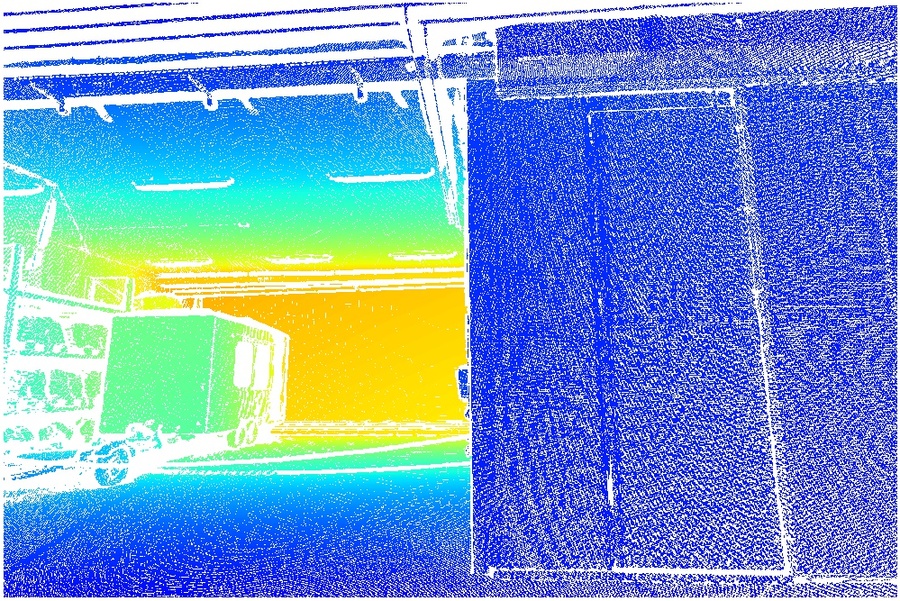} \\[0.5pt]
\includegraphics[width=0.14\textwidth]{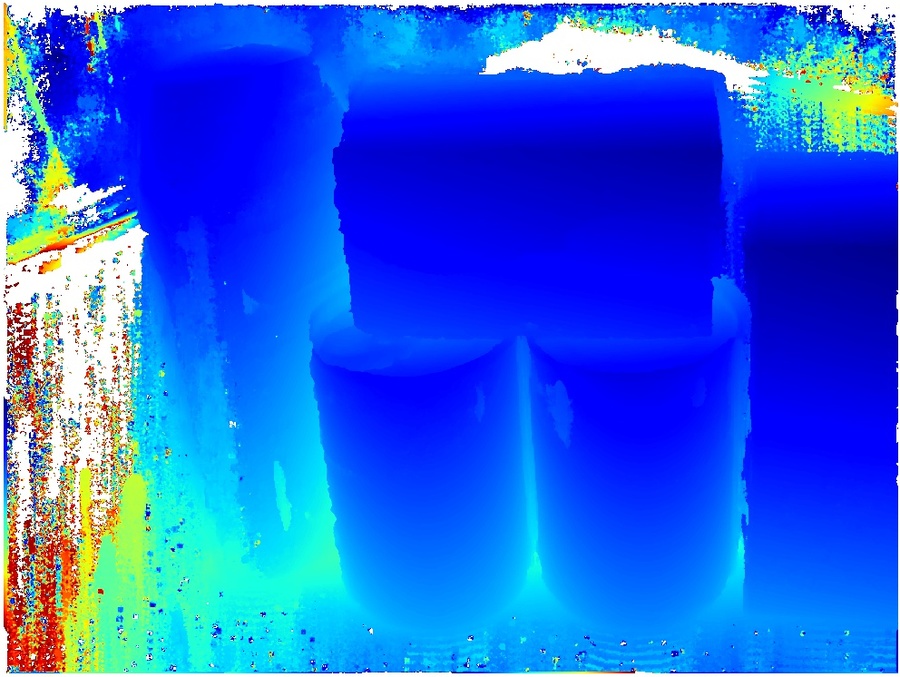} &
\includegraphics[width=0.16\textwidth]{pics/pipes_depth.jpg} &
\includegraphics[width=0.16\textwidth]{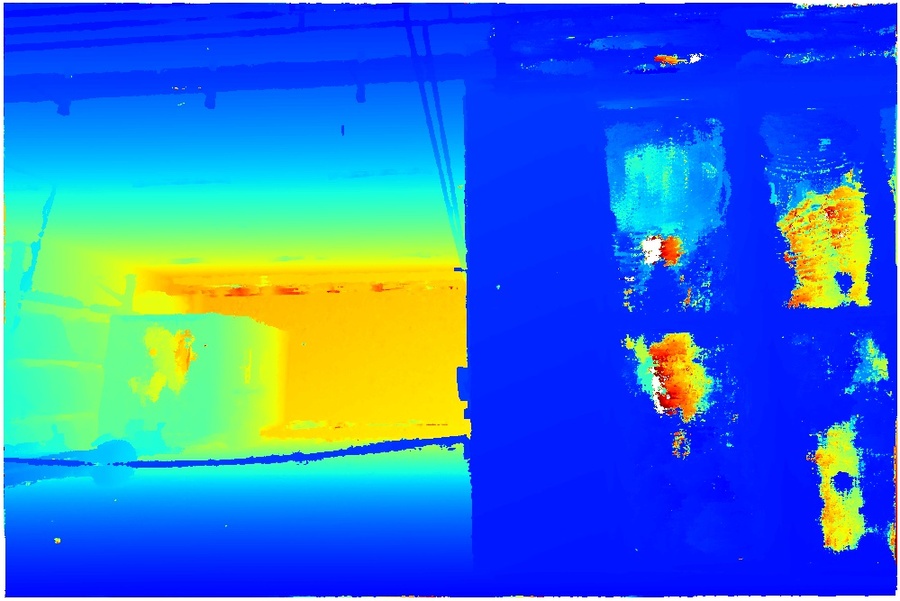} \\[0.5pt]
\includegraphics[width=0.14\textwidth]{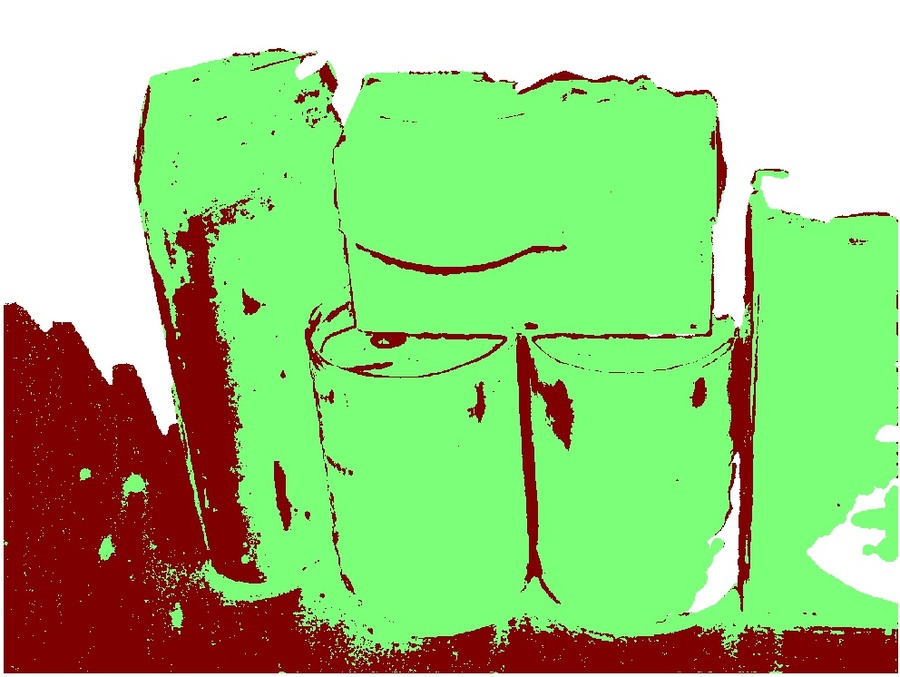} &
\includegraphics[width=0.16\textwidth]{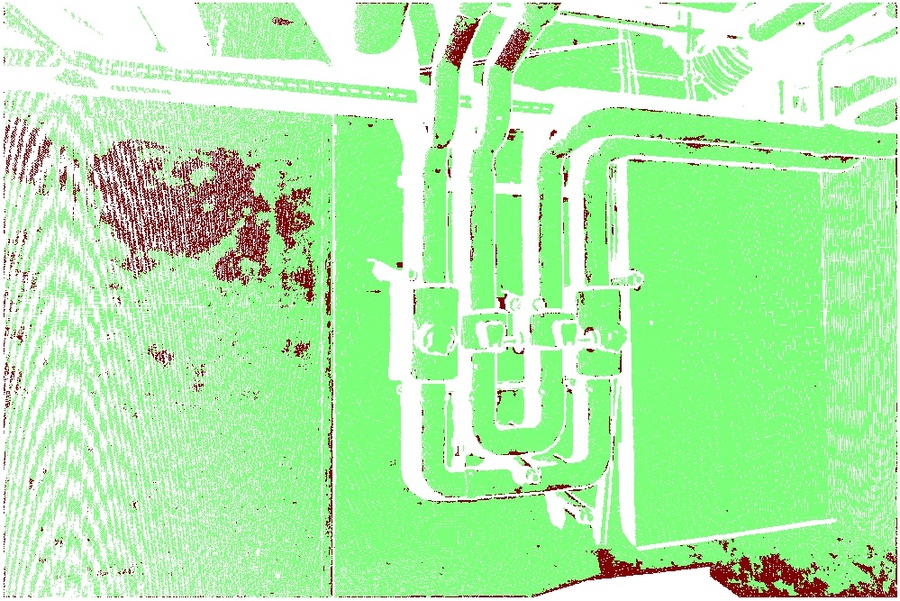} &
\includegraphics[width=0.16\textwidth]{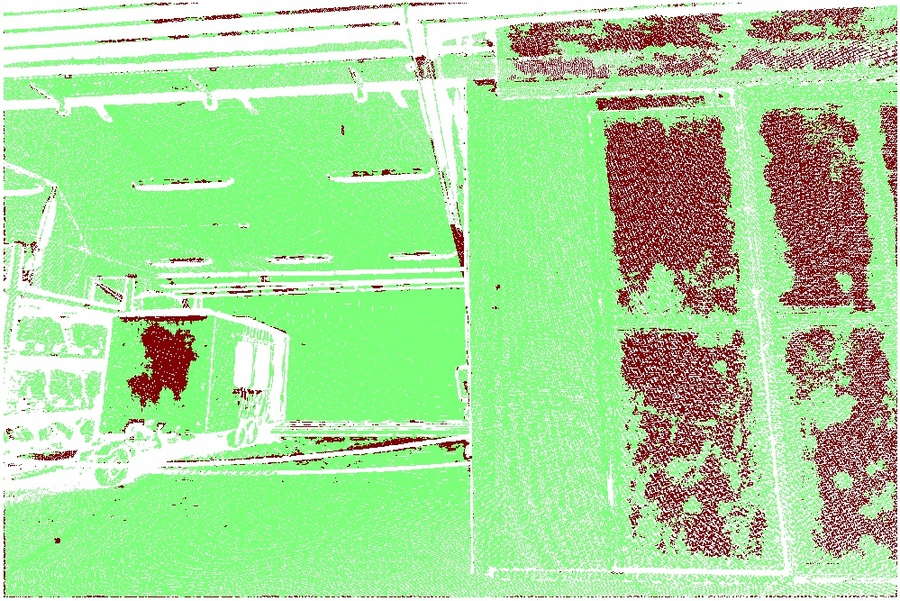}
\end{tabular}
\vspace{-6pt}
\caption{
 Examples from the DTU \cite{dtu:2016} (first column) and from the ETH3D training dataset \cite{schoeps2017cvpr} (other columns). From top to bottom: input image, ground truth, estimated depth map and ground truth label map. Missing ground truth depth and unreasonably far depth values are marked in white.
 In label maps, green denotes inliers, red outliers.
}
\label{fig:synth}
\end{figure}

Confidence prediction deals with the problem of estimating the probability that a given estimate
lies within a reasonable noise range.
We present a confidence prediction method capable of handling complex MVS scenarios using a DNN.
Our confidence prediction network is inspired by \textit{ConfNet} \cite{tosi:2018}, a network designed to estimate a pixel-wise confidence for stereo-derived disparity maps.

\textbf{Input:}
The ConfNet \cite{tosi:2018} input is the tensor obtained by concatenating the disparity map, whose confidence is sought, together with the corresponding RGB image.
In our MVS scenario, we deal with depth maps instead, which are not normalized, as the scale of a scene is generally not known a priori.
Moreover, the varying noise level in MVS-derived depth maps can be misleading for the network.
Therefore, rather than using the depth map as input to the network, we use the corresponding normal map, encoded in polar coordinates inside a two channel tensor.
Finally, since our MVS algorithm performs consistency checks employing the geometric and photometric error of depth measurements, we store the number of successfully matched images for every pixel and generate a counter map as additional network input.
A counter map example is provided in Fig.~\ref{fig:normalcounter}.

\begin{figure}[!h]
\centering
\includegraphics[width=0.48\textwidth]{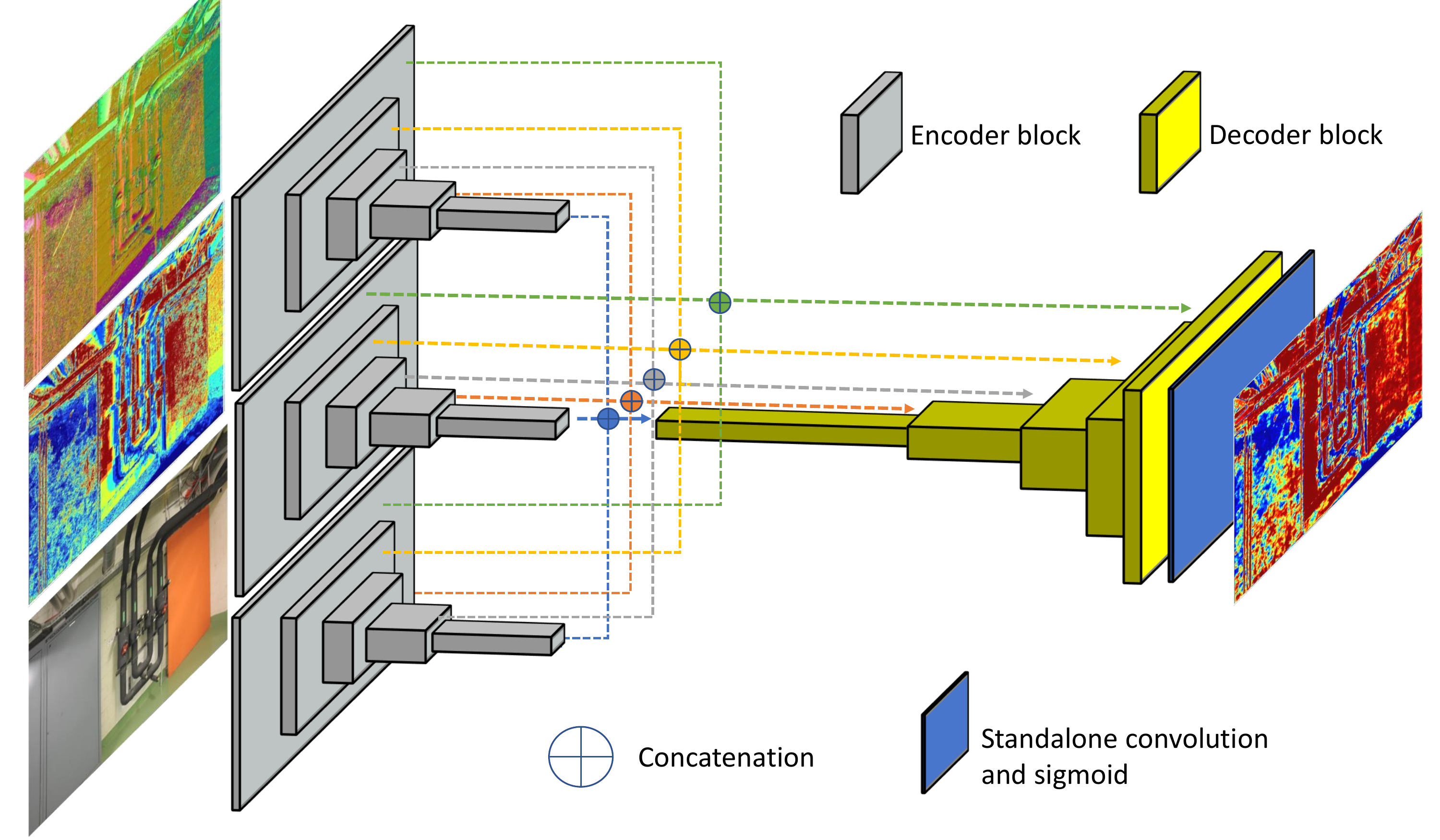} 
\vspace{-6pt}
\caption{Proposed deep confidence (DeepC) architecture.}
\label{fig:deepnetwork}
\end{figure}

\textbf{Architecture:}
Similarly to ConfNet, we resort to a U-Net \cite{RFB15a} architecture, but we adopt a middle fusion strategy.
In particular, we use a separate encoder for each one of our three inputs, namely the RGB image, the normal and the counter map, and we concatenate the resulting feature maps at each level before we pass them to the same level at the decoder side.
The architecture is depicted in Fig.~\ref{fig:deepnetwork}.
The encoder includes 4 blocks, each one consisting of two sub-blocks with 2D convolution, Batch Normalization and ReLU, followed by a single down-sampling by a factor of $2$.
The first encoder block outputs 32 feature maps; each subsequent block doubles their number.
The decoder is symmetric, however it employs only one sub-block.
Finally, the U-Net feature maps at the decoder output, which share the same spatial dimensions of the inputs, are fed to a block consisting of a 2D convolution followed by a sigmoid activation, which outputs the desired confidence map.

\textbf{Loss:}
ConfNet was originally trained on the Middlebury \cite{schar:2014} and KITTI \cite{Menze2015CVPR} datasets, for which the ground truth binary confidence maps can be generated easily by fixing a maximum disparity error, e.g., one pixel, and thresholding the input disparity maps accordingly.
In contrast, we work in the MVS scenario, with depth maps derived from multiple viewpoints.
As a consequence, fixing a noise level in meters is not possible, due to the varying baselines, focal lengths and distances to the scene.
Therefore, we project each pixel in the reference camera to all the corresponding source images, using the MVS-derived depth values whose confidence is sought.
We repeat the same procedure using the ground truth depth map instead, then we measure the Euclidean distance between the two obtained projections, in pixels.
If the distance is below a given threshold in at least one of the source images associated to the considered pixel, then the pixel is marked as an inlier, as an outlier otherwise.
We refer to the resulting binary map as a label map.
Label map examples are depicted in Fig.~\ref{fig:synth}.

\begin{figure*}[!ht]
\centering
\setlength{\tabcolsep}{1.0pt}
\renewcommand{\arraystretch}{0.1}
\begin{tabular}{cccc}
\includegraphics[width=0.24\textwidth]{pics/pipes_depth.jpg} &
\includegraphics[width=0.24\textwidth]{pics/pipes_normal.jpg} &
\includegraphics[width=0.24\textwidth]{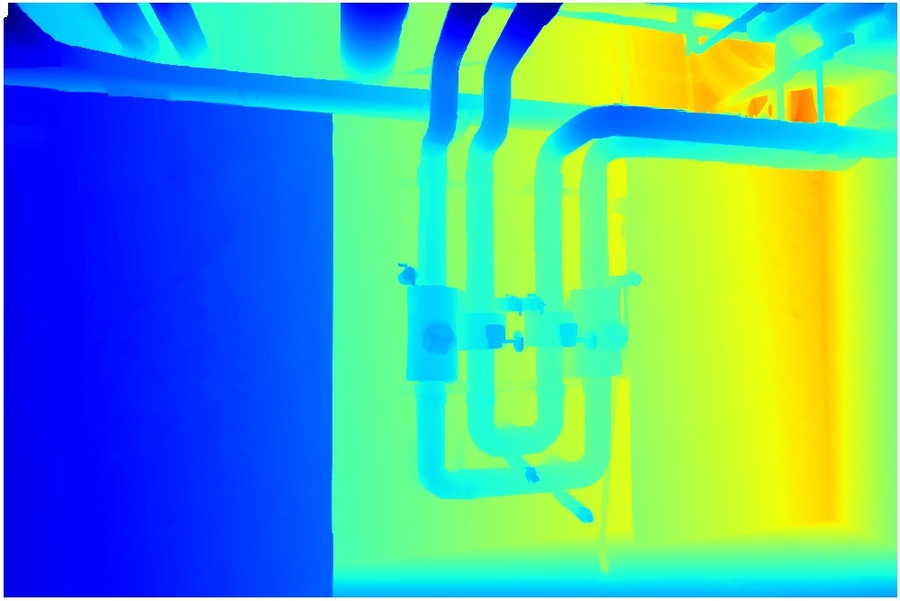} &
\includegraphics[width=0.24\textwidth]{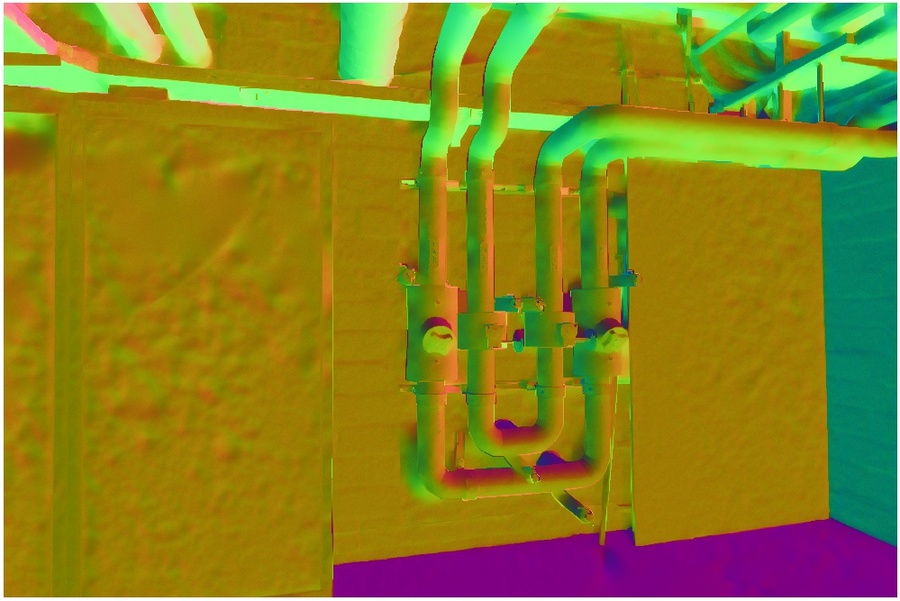}
\end{tabular}
\vspace{-6pt}
\caption{
{Original (left) and refined (right) depth and normal maps.}}
\label{fig:depth_and_normal_refined}
\end{figure*}

Finally, differently for ConfNet and traditional confidence prediction networks, we do not employ a Binary Cross Entropy loss, as it did not provide satisfying results with our unbalanced ground truth data.
The unbalanced data, with far more inlier labels than outlier ones, as it can be observed in Fig.~\ref{fig:synth}, can bias the network towards positive predictions.
Therefore, in a regression fashion, we adopt a loss based on the $\ell_2$--norm, where inliers and outliers are treated separately:
\begin{equation} \label{eq:loss}
    L \left( c, c_{gt} \right) \> = \tfrac{1}{|c_{gt}^{+}|} \> \lVert c^{+} - c_{gt}^{+} \rVert_2 \> + \tfrac{1}{|c_{gt}^{-}|} \> \lVert c^{-} - c_{gt}^{-} \rVert_2 \> ,
\end{equation}
with the vectors $c_{gt}^{+}$ and $c_{gt}^{-}$ gathering the inliers and outlier pixels labels, respectively, in the ground truth label map $c_{gt}$, the vectors $c^{+}$ and $c^{-}$ gathering the corresponding pixels in the predicted confidence map $c$ and $| \cdot |$ indicating the number of elements in a vector.
Having a loss term for each class, each one with an independent normalization, permits improved training on unbalanced data.

\section{Applications}

In the following we describe how the proposed learned confidence maps can be integrated in a 3D reconstruction pipeline to improve {the point cloud} quality.

\subsection{Filtering}

MVS-derived dense depth maps exhibit outliers caused by wrongly matched patches, hence
it is a common practice to filter them, pixel-by-pixel, within a post-processing step.
State-of-the-art MVS methods employ filtering, which considers the re-projection error from the reference image to the source ones and the photometric consistency \cite{schoenberger:2016mvs}.
In absence of a sufficient number of source images fulfilling the geometric and photometric requirements at a given pixel, the corresponding depth is filtered out.
In contrast, we provide that information, i.e., the counter map, as an additional input to our confidence prediction network.
Then, the depth map is filtered based on our estimated confidence map.

\subsection{Depth Refinement} \label{sec:refinement}

While filtering represents a straightforward approach to improve the accuracy of the final point cloud, it can increase its sparsity, as pixels with inaccurate depths are simply dropped.
Instead, depth refinement methods attempt to correct those pixels, which typically results in denser point clouds.
In general, refinement is more computationally demanding than filtering, but it also relies on confidences.

We consider the refinement method in \cite{rossi_cvpr_2019}, which leverages the piece-wise planarity characterizing most human made environments.
In particular, if the point $P_{j} \in \mathbb{R}^{3}$ belongs to the same plane of the point $P_{i} \in \mathbb{R}^{3}$ in the scene, with $i$ and $j \in \mathbb{R}^{2}$ the pixel coordinates where the two points are imaged in the camera, then the inverse depth $d$ at $j$ is in a planar relation with the inverse depth at $i$:
\begin{equation} \label{eq:plane_equation}
    d_{j} \> = \> d_{i} \> + \> u_{i}^{\top} \left( j - i \right),
\end{equation}
where $u_{i} \in \mathbb{R}^{2}$ is a vector defining the plane orientation at $i$.
Therefore, the authors in \cite{rossi_cvpr_2019} propose to refine a depth map $\bar{z}$ by enforcing that the refined inverse depth map $d = 1 / z$ is piece-wise planar.

The refinement of $\bar{d} = 1 / \bar{z}$ is cast into the minimization of a cost function comprising two terms:
\begin{equation*}
    \sum_{i} \left| \, d_{i} - \bar{d}_{i} \right| c_{i} \> + \>
    \lambda \> g \left( d, u \right),
\end{equation*}
where $\lambda \in \mathbb{R}_{\geq 0}$ balances them.
The left term penalizes those solutions deviating from the input $\bar{d}$ in those areas where this is considered as reliable by our confidence map $c$.
The right term $g(\cdot)$ is a regularization and promotes piece-wise planar inverse depth maps, in particular, it promotes the relation in Eq.\eqref{eq:plane_equation}.
The minimization is carried out with respect to both $d$ and $u$, therefore the method provides both the refined depth map and the corresponding normal map.
An example is provided in Figure~\ref{fig:depth_and_normal_refined}. We refer to \cite{rossi_cvpr_2019} for more details.

\section{Experiments}

In this section we analyze and test our proposed confidence prediction method and our overall MVS pipeline.
We trained our network jointly on the training datasets of ETH3D \cite{schoeps2017cvpr} high-resolution (high-res) and DTU \cite{dtu:2016}.
Four sequences were separated from the ETH3D high-res training dataset to be used as validation set.
We adopted ETH3D sequences for validation as our focus is real-world data and DTU provides images in laboratory setting only.
Finally, we tested our 3D reconstruction pipeline on the test datasets of ETH3D high-res and low-resolution (low-res) and Tanks and Temples \cite{Knapitsch:2017}.
All network configurations are trained for $100$ epochs with \textit{ADAM} \cite{Adam:2014} solver and a learning rate of $10e-1$.

\subsection{MVS} \label{sec:mvs}

We evaluate our \textit{vanilla MVS pipeline} (i.e., neither outlier filtering nor depth map refinement) on both the high and low-res training datasets of the ETH3D benchmark.
The benchmark provides scenes with ground truth laser scan point clouds, which are used to evaluate our reconstruction on an F$_{1}$ score metric computed as the harmonic mean of a \textit{completeness} and an \textit{accuracy} term \cite{schoeps2017cvpr}.
In our experiments we down sample the input images to half of their original resolution, perform $8$ red-black iterations of our MVS algorithm and adopt COLMAP \cite{schoenberger:2016mvs} depth fusion with the parameters of \cite{Xu:arxiv:2019} to allow a direct comparison and ablation study.
Table~\ref{tab:ablation_mvs} reports the results of our vanilla MVS pipeline (\textit{ours}) against the ones of ACMM \cite{Xu:arxiv:2019} and COLMAP \cite{schoenberger:2016mvs}, as our MVS algorithm is based on features from both these pipelines.
It can be observed that our pipeline achieves a better F$_{1}$ score than both ACMM and COLMAP.
The improvement is due to both an increase in the completeness score and competitive results in terms of accuracy.
The plane-based depth propagation \cite{schoenberger:2016mvs} is the main contributor to this improvement, as highly varying depth estimates along surfaces can now be propagated more effectively.
This can be observed by comparing \textit{ours} with \textit{ours without plane prop.} in the high-res side of Table~\ref{tab:ablation_mvs}.
On the low-res dataset there is a decrease in the F$_{1}$ score for \textit{ours} compared to \textit{ours without plane prop.} as the relative increase in completeness is smaller compared to high-res.
This is because low-res provides more redundant frames which already compensate for the lack in completeness of some scene regions.
In Table~\ref{tab:ablation_mvs} we also analyze the impact of our sampling pattern (\textit{ours}) against our same MVS algorithm with the ACMM sampling instead (\textit{ours with ACMM}): we observe that our proposed sampling pattern leads to an F$_{1}$ score improvement, due to an increased accuracy.

\begin{table}[!ht]
\centering
\small
\begin{tabular}{| c || p{1cm} | p{1cm} | p{1cm} || p{1cm} | p{1cm} | p{1cm} | }
\hline
\multirow{2}{*}{Method} & \multicolumn{3}{c||}{ETH3D high-res train} \\
 & F$_{1}$ & compl. & acc.  \\ \hline
COLMAP & 67.66  & 55.13 & \textbf{91.85}  \\
\hline
ACMM & 78.86 & 70.42 & 90.67  \\
\hline
ours & \textbf{83.62} & \textbf{83.25} & 84.17   \\
\hline
ours with ACMM pattern & 82.77 & 82.90 & 82.85    \\
\hline
ours without plane prop. & 81.23 & 78.96 & 83.97    \\
\hline
\end{tabular}

\vspace{-6pt}
\caption{
Comparison of our vanilla MVS pipeline (ours) to its variants, with ACMM sampling pattern and without plane propagation, as well as to COLMAP \cite{schoenberger:2016mvs} and ACMM \cite{Xu:arxiv:2019} on the ETH3D \cite{schoeps2017cvpr} training datasets.
}
\label{tab:ablation_mvs}
\end{table}
\vspace{-0.5em}

\subsection{Confidence Prediction}

\begin{figure*}[!ht]
\centering
\setlength{\tabcolsep}{0.5pt}
\renewcommand{\arraystretch}{0.1}
\begin{tabular}{cccc}
\includegraphics[width=0.32\textwidth]{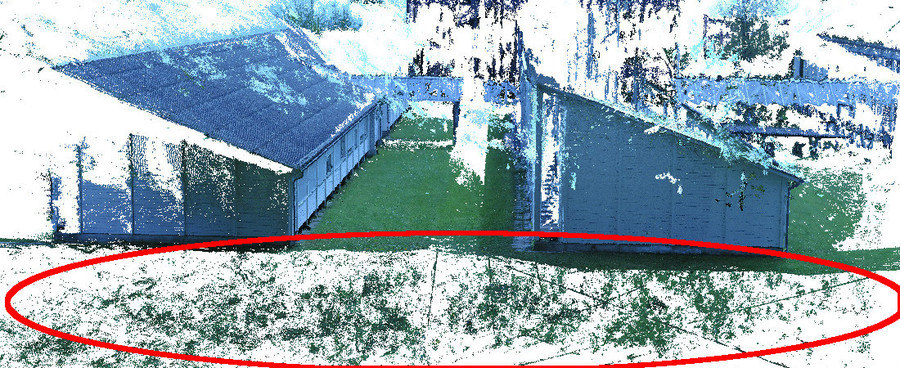} &
\includegraphics[width=0.32\textwidth]{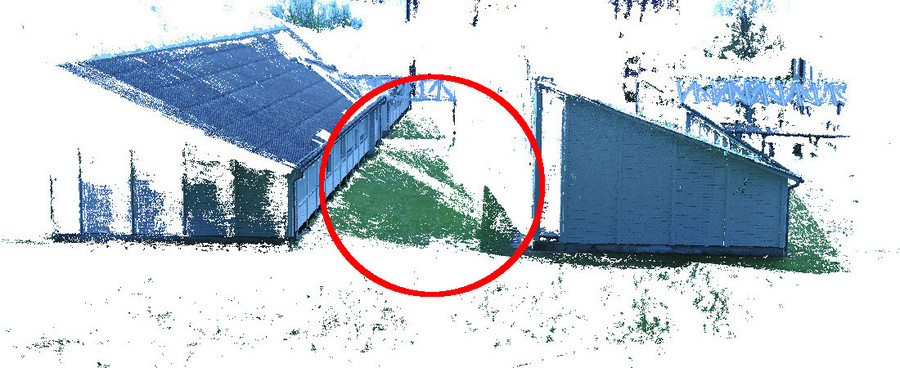}
\includegraphics[width=0.32\textwidth]{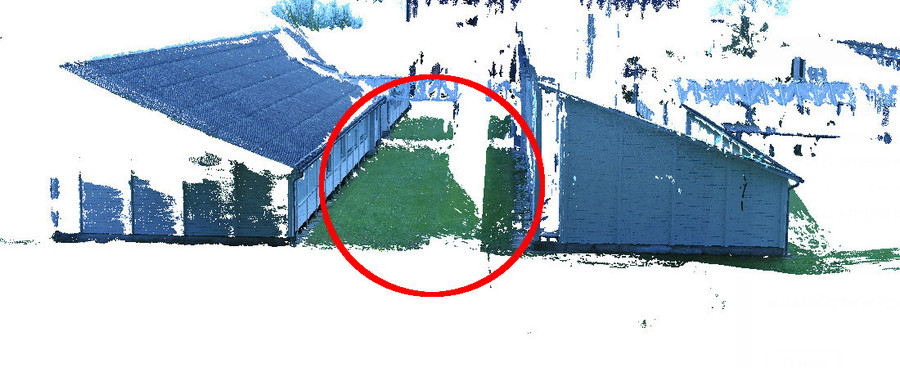}\\
\end{tabular}
\vspace{-6pt}
\caption{
\texttt{meadow} point clouds from ETH3D \cite{schoeps2017cvpr}. { Left to right: our vanilla MVS pipeline (\textit{ours}) and its versions with the 2 supporting view filtering (\textit{ours+mask2}) \cite{schoenberger:2016mvs,Xu:arxiv:2019} and confidence-based filtering (\textit{ours+conf$_{0.3}$}).
The vanilla MVS pipeline has the best F$_{1}$ score but needs improved outlier filtering.}}
\label{fig:filtering}
\end{figure*}

Confidence prediction targets a separation between correct and incorrect measurements.
As described in Sec.~\ref{sec:confpred}, we generated pixel-wise ground truth labels for each estimated depth map
and used $2$ pixels as the maximum projection distance in our experiments.

The effectiveness of confidence prediction methods is generally measured using the Receiver Operating Characteristic (ROC) curve.
The Area Under Curve (AUC) measures the separability of the two considered ROC curves, hence the quality of the predicted confidences.
Our confidence prediction algorithm handles MVS-derived depth maps, therefore it cannot be compared directly with the other confidence estimation methods in the literature, as these are generally meant for the simpler stereo scenario.
We rather compare the confidence maps obtained with the loss in Eq.~\eqref{eq:loss} against those obtained with the most common Binary Cross Entropy (BCE) loss and its weighted version, to handle the unbalanced training data.
In particular, for the weighted BCE we scaled the contribution of negative values by a factor of $3$, as we empirically found it to deliver the best results.
Our loss function provides the best AUC, equal to $96.42$, while BCE and its weighted version perform worse, providing $95.28$ and $95.08$, respectively.

Another major contribution of our work, as described in Sec.~\ref{sec:confpred}, is the proposal of a novel multi-channel input comprising the RGB image, the normal and the counter map, with features extracted separately in a middle fusion fashion.
We investigated the adoption of an early fusion strategy, similarly to ConfNet.
The resulting AUC is $95.22$, which shows a decrease of $1.22$ compared to our middle fusion with AUC $96.42$ and confirms the benefits our choice.

We conclude by observing that the normal map input permits the prediction of confidences for generic scenes, as it is scale independent, while the counter map input makes the network robust against complex MVS configurations.

\subsection{Filtering} \label{sec:filtering}

Here we analyse the benefits of complementing our vanilla MVS pipeline with outlier filtering.
Regarding COLMAP fusion parameters, we set $20^{\circ}$ as the maximum normal difference, $1$ pixel as the maximum re-projection error and $2$ as the minimum number of supporting views for 3D consistency.
For the dense sequences of the ETH3D low-res dataset instead, the maximum normal difference is quartered and the minimum number of supporting views is set to $3$, as they exhibit high redundancy.
In order to filter outliers, state-of-the-art MVS methods estimate both a geometric re-projection error and a photo-consistency measure \cite{schoenberger:2016mvs,Xu:arxiv:2019} for each reference image pixel and reject those with an insufficient number of supporting source images.
Table~\ref{tab:conffilter} compares our vanilla MVS pipeline results (\textit{ours}) to those obtained by adding the aforementioned filtering with the minimum number of supporting images set to one (\textit{mask1}) and two (\textit{mask2}), which are common numbers for datasets with a limited number of images like the ETH3D datasets.

Our confidence prediction network can implicitly merge the number of supporting views per pixel, i.e., the counter map, with other information like the level of texturedness of the RGB image and the level of depth noise suggested by the normal map.
Therefore, in Table~\ref{tab:conffilter} we report the results of our confidence-based filtering approach (\textit{ours+conf}), which rejects those pixels with a confidence smaller than a given threshold: $0.05$, $0.1$, $0.3$ and $0.5$.

\begin{table}[!h]
\centering
\small
\begin{tabular}{| p{2.8cm} || p{1.1cm} | p{1.1cm} | p{1.1cm} ||}
\hline
\multirow{2}{*}{Method} & \multicolumn{3}{c||}{ETH3D high-res train}  \\
 & F$_{1}$ & compl. & acc. \\ \hline
ours  & \textbf{84.52}    & \textbf{83.40}   & 85.87  \\ 
\hline
ours+conf$_{0.05}$ & \underline{84.29} & \underline{81.80} & 87.29 \\ 
ours+conf$_{0.1}$  & 83.64 & 79.95 & 88.26    \\ 
ours+conf$_{0.3}$  & 81.57 & 75.53 & 89.87    \\ 
ours+conf$_{0.5}$  & 79.72 & 72.29 & \underline{\textbf{90.71}}    \\ 
ours+mask1       & 84.08 & 81.44 & 87.27 \\ 
ours+mask2       & 79.95 & 73.22 & 89.5 \\ 
\hline
\end{tabular}

\vspace{-6pt}
\caption{
{Comparison of filtering methods on the \textit{ETH3D} \cite{schoeps2017cvpr} training datasets.}
Our confidence-based filtering (\textit{conf}) {has better F$_{1}$ scores compared} to mask filtering  \cite{schoenberger:2016mvs,Xu:arxiv:2019}.
The vanilla MVS pipeline, hence without filtering, is denoted as (\textit{ours}).
The absolute best scores are in bold while the best filtering scores are underlined.
}
\label{tab:conffilter}
\end{table}

In general, filtering improves the point cloud accuracy, as it minimizes the number of outliers.
However, for the $2$ supporting view (\textit{ours+mask2}) increased accuracy is paid by a dramatic decrease of the F$_{1}$ score.
Instead, our confidence-based filtering (\textit{ours+conf}) permits a better trade off between accuracy and F$_{1}$ score, as the threshold is continuous.
Compared to our vanilla MVS pipeline, their F$_{1}$ scores decrease slightly, but they lead to point clouds with a higher visual quality.
This can be appreciated in Fig.~\ref{fig:filtering}, where the point cloud resulting from the confidence-based filtering (\textit{ours+conf$_{0.3}$}) is characterized by a higher visual quality.
Finally, we observe that the vanilla MVS pipeline has the best F$_{1}$ score because outliers below the ground are not considered in the benchmark.

\subsection{Depth refinement} \label{sec:optimization}

In addition to outlier filtering, our predicted confidence can be used for a refinement of the depth maps from our vanilla MVS pipeline.
As described in Sec.~\ref{sec:refinement}, we used our confidence map within a global-optimization-based depth refinement framework: this regularizes depths and normals strongly in those areas marked as low confident, while it preserves fine details in high confident ones.
We adopted the same parameters proposed in \cite{rossi_cvpr_2019}.
The refined depth and normal maps were subsequently fused with the same method as the filtered maps in Sec.~\ref{sec:filtering}, besides that the minimum normal consistency was set to $5^{\circ}$, as the refined normal maps have a higher accuracy (see Fig.~\ref{fig:depth_and_normal_refined}).
For outlier filtering we trained our confidence map also for the refined depth maps and evaluated the results with filtering threshold $0.05$.
Table~\ref{tab:optimizer} shows the results for the ETH3D high-res training dataset.
The numerical improvements results especially from closed holes in the depth maps.
As demonstrated in Fig.~\ref{fig:deliveryarea}, the framework is also suitable for the refinement of thin structures, which unfortunately do not influence the numerical evaluation significantly.

\begin{table}[!h]
\centering
\small
\begin{tabular}{| p{3.2cm} || p{1cm} | p{1cm} | p{1cm} || p{1cm} | p{1cm} | p{1cm} |}
\hline
\multirow{2}{*}{Method} &   \multicolumn{3}{c||}{ETH3D high-res train}  \\
 & F$_{1}$             & compl          & acc.               \\ \hline
ours        & 84.52          & 83.40          & 85.87              \\
ours refined    & 84.80 & \textbf{81.00} & 89.45      \\
ours refined + conf$_{0.05}$    & \textbf{84.84}  & 80.42  & \textbf{90.31}       \\
\hline
\end{tabular}\\
\vspace{-6pt}
\caption{Our vanilla MVS pipeline, its version enhanced with refinement and refinement plus confidence-based filtering on the ETH3D \cite{schoeps2017cvpr} training datasets.}
\label{tab:optimizer}
\end{table}
\vspace{1.5em}

\subsection{Final Evaluation} \label{subsec:finaleval}

\begin{figure*}[!t]
\centering
\includegraphics[width=0.31\textwidth]{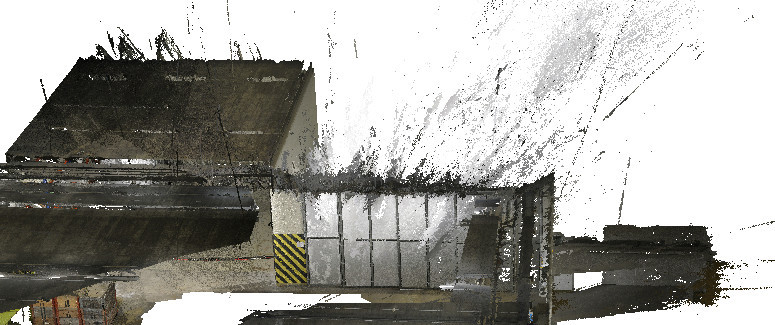} \,
\includegraphics[width=0.31\textwidth]{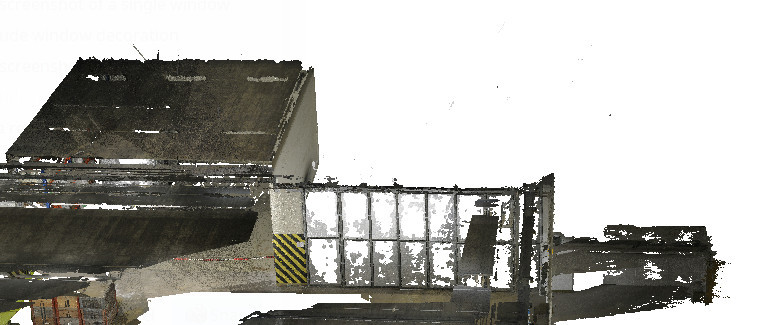} \,
\includegraphics[width=0.31\textwidth]{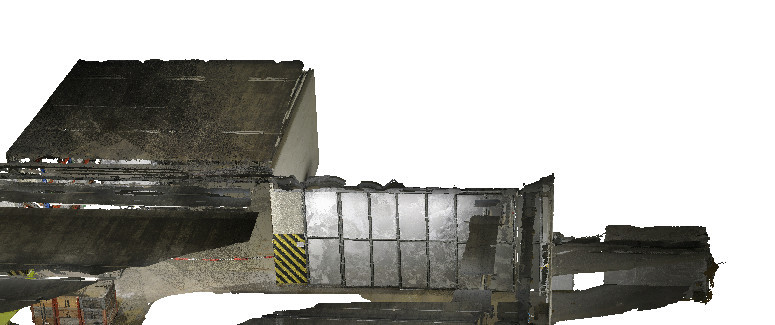} \\[2pt]
\includegraphics[width=0.31\textwidth]{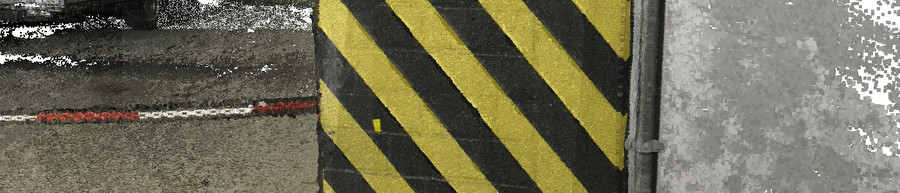} \,
\includegraphics[width=0.31\textwidth]{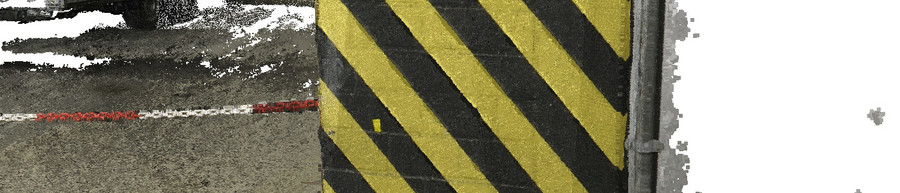} \,
\includegraphics[width=0.31\textwidth]{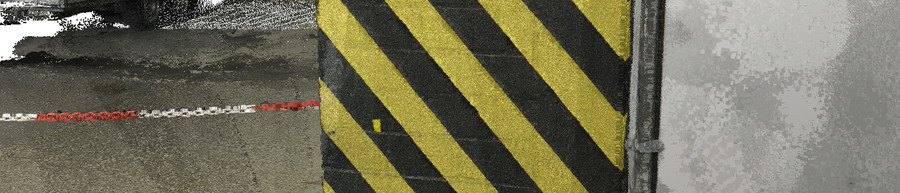}
\vspace{-6pt}
\caption{
\texttt{delivery area} point clouds from ETH3D \cite{schoeps2017cvpr}.
{Left: our vanilla MVS pipeline, center: ours with deep-confidence-based filtering (\textit{ours+conf$_{0.3}$}), right: ours  with refinement (ours refined + conf$_{0.05}$).
The refinement permits both the filtering of outliers and the filling of holes, while preserving thin structures as the red-white chain detail (bottom row).}
}
\label{fig:deliveryarea}
\end{figure*}

Firstly, we tested our proposed method with subsequent filtering and refining on the \textit{ETH3D} test datasets \cite{schoeps2017cvpr}.
We used the same parameters for all datasets as described in Sec.~\ref{sec:filtering} and \ref{sec:optimization}.
As the refinement method generates complete depth maps, geometric consistency can lack in the filtering of outliers in sky areas.
Therefore, we enable a sky filtering method \cite{kuhn:gcpr:2019} in order to minimize visual artifacts.
The sky filtering influences the ETH3D numerical evaluation only marginally ($0.01\%$ on the high-res F$_{1}$ score), but gives visually better results.

Our pipeline including the depth refinement method is effective but computationally complex, while the confidence-based filtering permits an efficient processing.
Therefore, we name our pipelines employing the refinement and the filtering as \textit{DeepC-MVS} and \textit{DeepC-MVS$_{\text{fast}}$} (or \textit{DeepC-MVS$_{\text{f}}$}), respectively.

\paragraph{ETH3D}
Table~\ref{tab:finaleth3d} shows a direct numerical comparison against the leading methods on the high-res and low-res datasets.
Both our pipelines outperform the current state-of-the-art, at the time of writing.
In particular, compared to ACMM, which we consider as the baseline method, our improvement on the ETH3D high-res data is $>6\%$.
From Fig.~\ref{fig:depth_and_normal_refined} and \ref{fig:deliveryarea} it is clear that the refinement improves significantly the flat surfaces, due to its piece-wise planarity bias, compared to the filtering which instead tends to sparsify the point cloud.

\begin{table}[!h]
\centering
\small
\begin{tabular}{| p{2.2cm} | p{0.8cm} | p{0.8cm} | p{0.8cm} | p{0.8cm} |}
\hline
\multirow{2}{*}{Method} & 
\multicolumn{2}{c|}{high-res} & 
\multicolumn{2}{c|}{low-res} \\ 
& train & test & train & test \\ 
\hline
DeepC-MVS & \textbf{84.81} & \textbf{87.08}         & 61.99          & \textbf{62.37} \\
$\text{DeepC-MVS}_{\text{f}}$    & 84.27 & 86.91 &  \textbf{62.57} & 62.24 \\
ACMM \cite{Xu:arxiv:2019}  & 78.86          & 80.78          & 55.12          & 55.01 \\  
PCF-MVS \cite{kuhn:gcpr:2019} & 79.42          & 79.29          & 57.32          & 57.06 \\ 
TAPA-MVS \cite{Romanoni:arxiv:2019} & 77.69          & 79.15          & 55.13          & 58.67 \\ 
LTVRE \cite{kuhn:ijcv:2017}   & 61.82          & 76.25          & 53.25          & 53.52 \\  
COLMAP  \cite{schoenberger:2016mvs}  & 67.66          & 73.01          & 49.91          & 52.32 \\  
P-MVSNet \cite{Luo_2019_ICCV} & n/a            & n/a            & n/a            & 44.46 \\
\hline
\end{tabular}
\vspace{-6pt}
\caption{The individual rows show the results (F$_{1}$ score $[\%]$) of the currently leading methods on ETH3D \cite{schoeps2017cvpr}.}
\label{tab:finaleth3d}
\end{table}

\paragraph{Tanks and Temples}
Additionally to the ETH3D benchmark \cite{schoeps2017cvpr}, the Tanks and Temples \cite{Knapitsch:2017} benchmark permits an evaluation of 3D point clouds derived from sets of images.
Its images are of relatively low resolution and come in a large number, similarly to the ETH3D low-res datasets, while in this article we focus on high-resolution image processing.
In addition, the Tanks and Temples benchmark evaluates the quality of the Structure from Motion (SfM) results, which is beyond the scope of this article.
For the sake of completeness, we discuss our evaluation in the following.
In contrast to ETH3D, no ground truth camera poses are given, therefore the results depend strongly on the employed SfM method.
When having inaccurately registered images, the improved completeness of our depth maps can have a negative influence on the overall F$_{1}$ score, as all points are marked as outliers.
We observed that the employed COLMAP SfM lacks in accurately registered images, especially for the advanced sequences.
Nonetheless, overall we achieve state-of-the-art results and have best scores on the intermediate sequences, as shown in Table~\ref{tab:finaltt}.
Compared to ACMM, we achieve higher scores in the \textit{precision} (accuracy) but lower score in the \textit{recall} (completeness).
This can be traced back to differing parameters in the 3D fusion, as ACMM parameters for Tanks and Temples are not available.
On our side, we use the same parameters as for the ETH3D low-res sequences, without tuning them for the Tanks and Temples dataset.

\begin{table}[!h]
\small
\begin{tabular}{| p{2.05cm} | p{0.55cm} p{0.55cm} p{0.55cm} | p{0.55cm} p{0.55cm} p{0.55cm} |}
\hline
\multirow{2}{*}{Method} & 
\multicolumn{3}{c|}{Intermediate} &  
\multicolumn{3}{c|}{Advanced} \\ 
& F$_{1}$ & Rec. & Prec. & F$_{1}$ & Rec. & Prec. \\  
\hline
DeepC-MVS & \textbf{59.79} & 61.21 & \textbf{59.11} & 34.54 & 31.30 & \textbf{40.68} \\  
$\text{DeepC-MVS}_{\text{f}}$ & 57.47 & \textbf{62.75} & 54.51 & 35.16 & 33.58 & 39.14  \\ 
ACMM \cite{Xu:arxiv:2019}  & 57.27 & 70.85 & 49.19 & 34.02 & \textbf{37.40} & 35.63 \\
CasMVSNet \cite{DBLP:journals/corr/abs-1912-06378} & 56.84 & 74.01 & 47.62 & 31.12 & 35.24 & 29.68 \\
P-MVSNet \cite{Luo_2019_ICCV} & 55.62 & 63.82 & 49.93 & & n/a &         \\
PCF-MVS \cite{kuhn:gcpr:2019} & 53.39 & 58.85 & 50.04 & \textbf{34.59} & 34.35 & 35.84 \\
ACMH \cite{Xu:arxiv:2019} & 54.82 & \textbf{77.54} & 43.44 & 33.73 & 37.40 & 34.50 \\
COLMAP \cite{schoenberger:2016mvs} & 42.14 & 44.48 & 43.16  & 27.24 & 23.96 & 33.65 \\
\hline
\end{tabular}
\vspace{-6pt}
\captionof{table}{
{Average F$_{1}$ score, recall and precision $[\%]$ at varying distances, as defined by \textit{Tanks and Temples} \cite{Knapitsch:2017} evaluation software.}
}
\label{tab:finaltt}
\end{table}

Tables~\ref{tab:finaleth3d} and \ref{tab:finaltt} also show comparisons with recent end-to-end deep-learning-based MVS pipelines \cite{Luo_2019_ICCV,DBLP:journals/corr/abs-1912-06378}.
A major disadvantage of these methods is their processing of large 3D cost volumes, which is computationally and memory intensive, and prevents their evaluation on high-resolution datasets like the ETH3D high-res.
In addition, their results on the ETH3D low-res are far below the state-of-the-art.
Instead, our method can process high resolution image sets, is less biased toward the training data, as it relies on training only for the confidence prediction sub-task, and achieves state-of-the-art results for all the considered datasets.

\section{Summary}

We presented
a confidence prediction network for depth maps derived from challenging Multi-View Stereo (MVS) configurations.
The predicted confidence maps were used for improved outlier filtering as well as for depth map refinement, thus leading to two 3D reconstruction pipelines.
The first, DeepC-MVS, is suitable for dense high quality reconstruction.
The second, DeepC-MVS$_{fast}$, is more suitable for those scenarios where a faster computation is required.
Tests on popular benchmarks showed that both our pipelines produce state-of-the-art 3D reconstructions, qualitatively and quantitatively, while being able to handle large and high resolution image sets.

{\small
\bibliographystyle{ieee}
\bibliography{refs}
}

\end{document}